\def\year{2018}\relax
\documentclass[letterpaper]{article} 
\usepackage{aaai18}  
\usepackage{times}  
\usepackage{helvet}  
\usepackage{courier}  
\usepackage{url}  
\usepackage{graphicx}  
\begin{document}
%

\title{Predicting Depression Severity by Multi-Modal Feature Engineering and Fusion} 

\newcommand*\samethanks[1][\value{footnote}]{\footnotemark[#1]}
 \author{Aven Samareh\thanks{Equal contributor} \textsuperscript{1}, Yan Jin \samethanks  \textsuperscript{1}, Zhangyang Wang \textsuperscript{2}, Xiangyu Chang \textsuperscript{3}, Shuai Huang\textsuperscript{1} \\\textsuperscript{1} Industrial \& Systems Engineering Department, University of Washington  \\ \textsuperscript{2} Department of Computer Science and Engineering, Texas A\&M University \\ \textsuperscript{3} Department of Information, Xi'an Jiaotong University \\
 \{asamareh, yanjin, shuaih\}@uw.edu, atlaswang@tamu.edu, xiangyuchang@gmail.com  
}

\maketitle
\begin{abstract}
We present our preliminary work to determine if patient's vocal acoustic, linguistic, and facial patterns could predict clinical ratings of depression severity, namely Patient Health Questionnaire depression scale (PHQ-8). We proposed a multi-modal fusion model that combines three different modalities: audio, video, and text features. By training over the AVEC2017 dataset, our proposed model outperforms each single-modality prediction model, and surpasses the dataset baseline with a nice margin.
\end{abstract}

\section{ Introduction }
Depression is a significant health concern worldwide, and its early-stage symptom monitoring, detection, and prediction are becoming crucial for us to mitigate this disease. With considerable attentions devoted to this field, traditional diagnosis and monitoring procedures usually rely on subjective measurements. It is desirable to develop more biomarkers that can be automatically extracted from objective measurements. Depression will leave recognizable markers in patient's vocal acoustic, linguistic, and facial patterns, all of which have demonstrated increasing promise on evaluating and predicting patient's mental condition in an unobtrusive way \cite{kachele2014fusion}. In this work, we aim to extend the existing body of related work and investigate the performances of each of the biomarker modalities (audio, linguistic, and facial) for the task of depression severity evaluation, and further boost our results by using a confidence based fusion mechanism to combine all three modalities. Experiments on the recently released AVEC 2017 \cite{AVEC2017} depression dataset have verified the promising performance of the proposed model.

\section{Feature Engineering}

The AVEC 2017 dataset includes audio and video recordings, as well as extensive questionnaire responses in text formats, collected from (nearly) real-world settings. We will next introduce how we developed feature engineering techniques based on given data and features in each modality. 

The original audio datasets were pre-extracted features using the COVAREP toolbox.  We further extracted descriptors: fundamental frequency (F0), voicing (VUV), normalized amplitude quotient (NAQ), quasi open quotient (QOQ), the first two harmonics of the differentiated glottal source spectrum (H1, H2), parabolic spectral parameter (PSP), maxima dispersion quotient (MDQ), spectral tilt/slope of wavelet responses (peak/slope), shape parameter of the Liljencrants-Fant model of the glottal pulse dynamic (Rd), Rd conf, Mel cepstral coefficient (MCEP 0-24), harmonic model and phase distortion mean (HMPDM 0-24) and deviations (HMPDD 0-12), and the first 3 formants. The top $10$ largest discrete cosine transformation (DCT) coefficients were computed for each descriptor to balance between information loss and efficiency. Delta and Delta-Delta features known as differential and acceleration coefficients were calculated as additional features to capture the spectral domain dynamic information. In addition, a series of statistical descriptors such as mean, median, std, peak-magnitude to rms ratio, were calculated. Overall, a total of 1425 audio features were extracted.

2D coordinates of 68 points on the face, estimated from raw video data were provided. To develop visual features from this data-limited setting, we chose stable regions between eyes and mouth due to minimal involvement in facial expression. We calculated the mean shape of 46 stable points not confounding with gender. The pairwise Euclidean distance between coordinates of the landmarks were calculated as well as the angles (in radians) between the points, resulting in 92 features. Finally, we split the facial landmarks into three groups of different regions: the left eye and left eyebrow, the right eye and right eyebrow, and the mouth. We calculated the difference between the coordinates of the landmarks and finally calculated the Euclidean distances ($\ell_2$-norm) between the points for each group, resulting in 41 features. Overall, we obtained 133 features.

The transcript file includes translated communication content between each participant and the animated virtual interviewer ‘Ellie’. Basic statistics of words or sentences from the transcription file including number of sentences over the duration, number of the words, ratio of number of the laughters over the number of words were calculated. The depression related words were identified from a dictionary of more than 200 words downloaded from online resources\footnote{\url{https://myvocabulary.com/word-list/depression-vocabulary/}}. The ratio of depression-related words over the total number of words over the duration was calculated.  

In addition, we introduced a new set of \textbf{text sentiment} features, obtained using the tool of AFINN sentiment analysis \cite{nielsen2011new}, that would represent the valence of the current text by comparing it to an exiting word list with known sentiment labels. The outcome of AFINN is an integer between minus five (negative) and plus five (positive), where negative and positive number number shows negative and positive positive sentiment subsequently. The mean, median, min, max, and standard deviation of the sentiment analysis outcomes (as a time series) were used. A total of 8 features were extracted. The new set of sentiment features was found to be highly helpful in experiments.

\section{Multi-Modal Fusion Framework}

\begin{figure}[ht]
\centering
\includegraphics[width=0.5\textwidth]{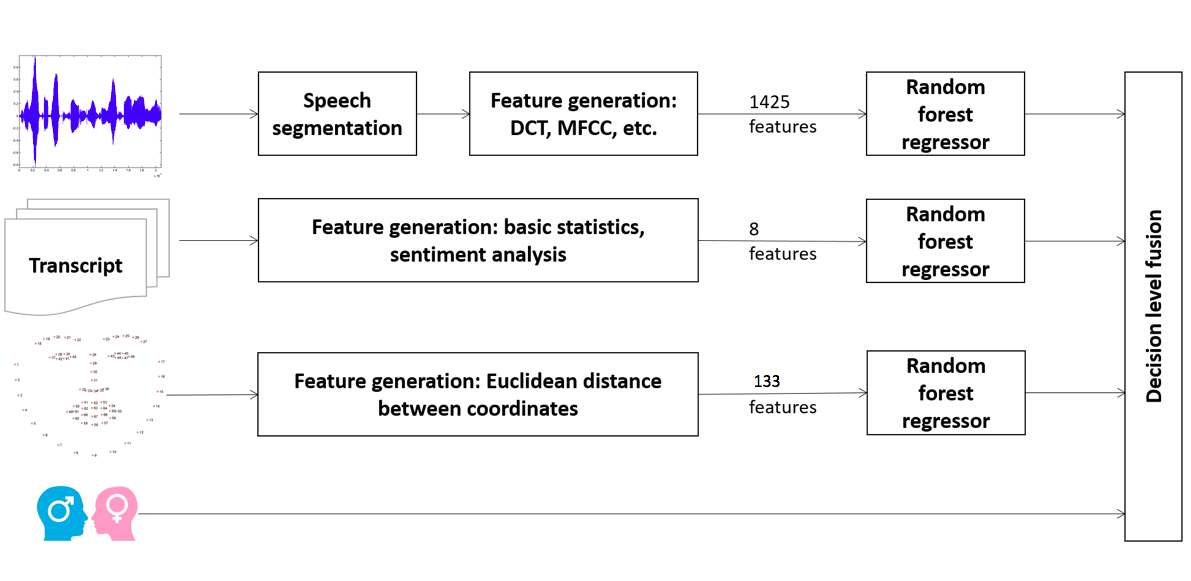}
\caption{Overview of the confidence based decision-level fusion method}
\label{fig:framework}
\end{figure}

We adopted an input-specific classifier for each modality, followed by a decision-level fusion module to predict the final result. In detail, for each modality biomarker we used a random forest to translate features into predictive scores, while these scores were further combined in a confidence based fusion method to make final prediction on the PHQ8. To fuse the modalities, we implemented a decision-level fusion method. Rather than simple averaging, we recognized that each modality itself might be noisy. Therefore, for each modality we calculated the standard deviation for the outcomes of all trees, defined as the modality-wise \textbf{confidence score}. After trying several different strategies, the \textit{winner-take-all} strategy, i.e., picking the single-modality prediction with the highest confidence score as the final result seems to be the most effective and reliable in our setting. In most cases, we observed that audio modality tends to dominate during the prediction. We conjectured that it implies the imbalanced (or say, complementary) informativeness of three modalities, and one modality often tends to dominate in each time of prediction. An overview of the confidence based decision-level fusion method is shown in Figure \ref{fig:framework}.

\section{Preliminary Result and Future Work }
Baseline scripts provided by AVEC have been made available in the data repositories where depression severity was computed using random forest regressor.
Table~\ref{tab:perf_fusion} reports the performance of the baseline and our model for development and training sets. For both models, we reported the performance of single modality and multi-modal fusion methods. Comparing to the baseline, confidence based fusion could achieve comparable or even marginally better performance than the baseline in terms of both RMSE and MAE.

\begin{table}[thp]
\caption{Performance comparison among single modality and confidence based fusion model}
\centering
\label{tab:perf_fusion}
\begin{tabular}{  c| c c | c c }
\hline
Feature used & \multicolumn{2}{c}{`development'} & \multicolumn{2}{c}{`train'} \\
& \textbf{RMSE} & MAE & RMSE & MAE \\
\hline \hline
\multicolumn{5}{c}{The baseline provided by AVEC organizer} \\
\hline 
Visual only & 7.13 & 5.88 & 5.42 & 5.29 \\
Audio only & 6.74 & 5.36 & 5.89 & 4.78 \\
Audio \& Video & 6.62 & 5.52 & 6.01 & 5.09 \\
\hline 
\multicolumn{5}{c}{Our model that doesn't include gender variable} \\
\hline
Visual only & 6.67 & 5.64 & 6.13 & 5.08 \\
Audio only & 5.45 & 4.52 & 5.21 & 4.26 \\
Text only & 5.59 & 4.78 & 5.29 & 4.47 \\
Fusion model & 5.17 & 4.47 & 4.68 & 4.31 \\
\hline
\multicolumn{5}{c}{Our model that includes the gender variable} \\
\hline
Visual only & 5.65 & 4.87 & 4.99 & 4.46 \\
Audio only & 5.11 & 4.69 & 4.84 & 4.23 \\
Text only & 5.51 & 4.87 & 5.13 & 4.28 \\
Fusion model & 4.81 & 4.06 & 4.23 & 3.89 \\
\hline
\end{tabular}
\end{table}

We plan to enhance our methodology in the following directions. First, to improve decision rules, we will use Rule ensemble models to exhaustively search interactions among features and scale up the high-dimensional feature space. In addition, we are interested to perform vowel formants analysis to allow a straightforward detection of high arousal emotions. Second, we found that with more relevant features refined, the overall performance could be improved (e.g.,  silence detection). Finally, we plan to implement our model to a more general clinical environment (e.g., routine patient-provider communication) to characterize social interactions to support clinicians in predicting depression severity. 

 \pdfoutput=1

\bibliography{formatting-instructions-latex-2018.bbl}
\bibliographystyle{aaai}
\end{document}